\documentclass[10pt,twocolumn,letterpaper]{article}

\usepackage{wacv}
\usepackage{times}
\usepackage{epsfig}
\usepackage{graphicx}
\usepackage{amsmath}
\usepackage{amssymb}
\usepackage[english,algoruled,linesnumbered,vlined]{algorithm2e}
\let\oldnl\nl% Store \nl in \oldnl
\newcommand{\nonl}{\renewcommand{\nl}{\let\nl\oldnl}}% Remove line number for one line

\usepackage{url}
\usepackage[hyperfootnotes=false]{hyperref}

\usepackage{wrapfig}
\usepackage{float}
\usepackage{tabu,multirow}
\restylefloat{table}
\usepackage{placeins}
\usepackage{booktabs}
\usepackage{footnote}
\usepackage{tablefootnote}
\usepackage{afterpage}

\usepackage[table,xcdraw]{xcolor}

\usepackage{mathtools}  
\mathtoolsset{showonlyrefs}

\usepackage[caption=false]{subfig} % for subfigures
\usepackage{siunitx} % used for degrees symbol

\def\mat#1{\mathchoice{\mbox{\boldmath$\displaystyle\tt#1$}}
{\mbox{\boldmath$\textstyle\tt#1$}}
{\mbox{\boldmath$\scriptstyle\tt#1$}}
{\mbox{\boldmath$\scriptscriptstyle\tt#1$}}}

\def\vec#1{\mathchoice{\mbox{\boldmath  $\displaystyle\bf#1$}}
{\mbox{\boldmath  $\textstyle\bf#1$}}
{\mbox{\boldmath  $\scriptstyle\bf#1$}}
{\mbox{\boldmath  $\scriptscriptstyle\bf#1$}}}

\newlength{\colwidth}
\newcommand{\todo}[1]{}
\renewcommand{\todo}[1]{{\color{red} TODO: {#1}}}

\usepackage{manyfoot}

\DeclareNewFootnote{A}
\DeclareNewFootnote{B}

\let\footnoteR\footnoteB
\let\footnote\footnoteA

\wacvfinalcopy % *** Uncomment this line for the final submission

\def\wacvPaperID{66} % *** Enter the wacv Paper ID here

\ifwacvfinal\fi
\begin{document}

%%%%%%%%% TITLE
\title{RGBD2lux: Dense light intensity estimation with an RGBD sensor} %Is 3D modeling of a real scene with a single \\camera sufficient for reliable light modeling? TOCHANGE}

% Authors at the same institution
%\author{First Author \hspace{2cm} Second Author \\
%Institution1\\
%{\tt\small firstauthor@i1.org}
%}
% Authors at different institutions
% \author{Theodore Tsesmelis \\
% Institution1\\
% {\tt\small firstauthor@i1.org}
% \and
% Irtiza Hasan \\
% Institution2\\
% {\tt\small secondauthor@i2.org}
% }

\author{Theodore Tsesmelis$^{1,2,3}$, Irtiza Hasan$^{3,1}$, Marco Cristani$^{2,3}$, Fabio Galasso$^{1, \dagger}$, Alessio Del Bue$^{2, \dagger}$\\
{\normalsize Corporate Innovation OSRAM GmbH$^{1}$, Istituto Italiano di Tecnologia (IIT)$^{2}$, University of Verona (UNIVR)$^{3}$}\\
% Institution1 address\\
{\tt\small t.tsesmelis@osram.com, irtiza.hasan@univr.it}}

\maketitle
\ifwacvfinal\thispagestyle{empty}\fi

\begin{abstract} \label{sec:abstract}
Lighting design and modelling %(the efficient and aesthetic placement of luminaires in a virtual or real scene) 
or industrial applications like luminaire planning and commissioning %(the luminaire's installation and evaluation process along to the scene's geometry and structure) 
rely heavily on time-consuming manual measurements or on physically coherent computational simulations. Regarding the latter, standard approaches are based on CAD modeling simulations and offline rendering, with long processing times and therefore inflexible workflows. Thus, in this paper we propose a computer vision based system to measure lighting with just a single RGBD camera. The proposed method uses both depth data and images from the sensor to provide a dense measure of light intensity in the field of view of the camera.   % examine whether a 3D modeling simulation based on camera-aided information could be used to accurately simulate the light distribution of a luminaire on the fly and to answer the question how sufficiently meaningful and reliable such a light modeling procedure could be. 
% We evaluate our system to the available ground truth data as well as to corresponding offline simulation outputs from a commercial light-planning software and the results shows a converge towards a positive conclusion. Thus, our work contributes by setting the first baseline for using a camera input as the only requirement to light measurement methods in a general indoor scenario and examines in practise whether such an input could be used to accurately estimate the light levels in the scene. Consequently, answering to the question about how sufficiently meaningful and reliable such a light modeling solution could be. %for light estimation and modeling, and opens a new applicability where the illumination modeling can be turned into an interactive process, allowing for real-time modifications and immediate feedback on the spatial illumination of a scene over time.
We evaluate our system on novel ground truth data and compare it to state-of-the-art commercial light-planning software. Our system provides improved performance, while being completely automated, given that the CAD model is extracted from the depth and the albedo estimated with the support of RGB images. To the best of our knowledge, this is the first automatic framework for the estimation of lighting in general indoor scenarios from RGBD input.
\end{abstract}
% \vspace{-5pt}
\section{Introduction} \label{sec:introduction}
\footnoteR{$^{\dagger}$These two authors contribute equally to the work.}
Most lighting systems around us are the result of a careful design by lighting professionals and architects. This is the case of offices, whereby the level of light on each desk is regulated by ISO standards %, retail shops (according to specific marketing strategies), 
and of industrial sites, whereby lighting means safety and well-being for the human operators. In recent years, most sectors have also changed to more agile strategies, \eg \ reconfiguring the position of office desks in due course according to the need, which is not reflected any more in the previously designed lighting system.

The industrial-driven process of designing and creating lighting systems relies heavily on computer graphic based simulations. These approaches have been given an increased scientific attention \cite{lin2013, schwarz2014, sorger2016} with modern systems where the light design process can be carried out in a completely interactive way using novel methods from the field of visual computing \cite{hilite,litemaker}. 
Besides such improvements, this process requires a strong manual intervention, using tools that are essentially custom-build computer graphics and computer-aided design (CAD) software \cite{dialux,agi32,relux}. All methods provide dense light intensity measurements through simulation, but they are based on simplified CAD models of a scene and they require to manually assign the reflectance properties of each structure and object present in the environment. 

On the other hand, the only alternative for measuring light is hardware-based and it relies on the use of luxmeters, a device that normally provide a point-to-point measurements (\ie sparse) of light intensity. This means that the operator needs to repeat the measurements in several positions of the environment, swaying for angular and spatial completeness and as a consequence being a very time and cost-expensive procedure.

In this paper, we propose an efficient solution to light measurement which uses commodity hardware (an RGBD sensor) and it provides dense pixel-level measurements of the light intensity in real environments. We leverage best features from  CAD-based simulation approaches, \ie to achieve a dense lighting measurement, but we bypass the need of user input of the CAD and reflectance models by means of a RGBD camera observing the scene. % the RGB (for the albedo) and depth (for the geometry) imaging. 
This results in an automatic procedure, which reaches the accuracy required by lighting designers.

\begin{figure*}[!ht]
	\begin{center}
	    \includegraphics[width=1\linewidth]{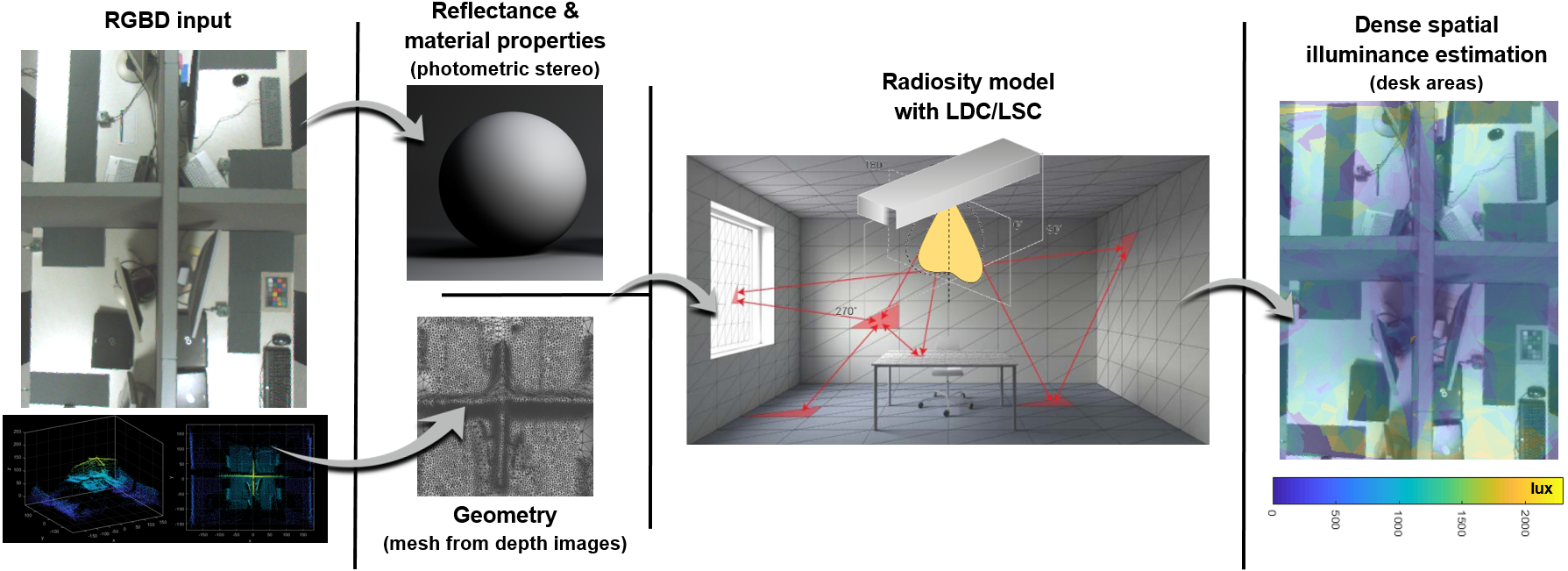}
	\end{center}
	\vspace{-5pt}
	\caption[]{Pipeline of our system. We first acquire the RGBD input from the camera system (\textit{left}), we then use the RGB images for extracting the albedo values of the surfaces based on a photometric stereo solution and the depth images for extracting the actual surfaces and the partial geometry of the scene (\textit{center}). Both are then fed into our radiosity model using known light sources distribution curves. Lastly we measure the estimated illumination over the visible scene. See the estimated illumination over the desk areas (\textit{right}).}\label{fig:teaser}
 	\vspace{-5pt}
\end{figure*}

Figure \ref{fig:teaser} shows a graphical description of our system (RGBD2lux). In general, the RGBD camera is installed so that to provide a top-view of the indoor environment\footnote{This is a practical solution since it provides the least occluded view of an indoor environment, however our method by design could work from any viewpoint.}. Given the depth information, we extract a partial mesh in the field of view of the camera. At the same time, for each patch of the mesh we estimate the reflectance as a scalar value (albedo) from multiple images with different light sources activated. Finally, the 3D mesh and reflectance are then used in a radiosity model that provides us with the overall estimation of the scene illuminance. The proposed system is then evaluated against ground truth readings from a set of luxmeters showing its potential as a new system for dense light measurement.

The rest of the paper is organized as follows. Section \ref{sec:previous_work} presents related work and the system introduced in this paper. Then Section \ref{sec:RGBD2Lux approach} describes step by step the system for estimating light intensity from RGBD images. In Section \ref{sec:exps} we introduce the benchmark and conduct experiments and ablation studies. Finally Section \ref{sec:conclusion} concludes the paper.

%\addtocounter{footnote}{-1}\footnotetext{Desk areas are considered of high importance in the lighting field.}
\section{Related Work} \label{sec:previous_work}

Measuring light is a problem that can be addressed by different fields. In the following we review three major topics in the literature related to light measurements. 

\noindent \textbf{Image processing.}  According to Cuttle \etal \cite{Cuttle2010} the lighting profession and lighting evaluation procedures are moving from the conventional illuminance-based towards the luminance-based. That means to move from assessing light incident on planes (\eg illuminance) towards the assessing light arriving at the eye (\ie luminance). Given this change, the works in \cite{cai2016luminance,choo2015,hiscocks2014measuring} take advantage of the emergence of camera-aided light measurement solutions. Hiscocks \etal \cite{hiscocks2014measuring} in his white paper provides an overall understanding of the luminance measuring procedure with a digital camera. Choo \etal \cite{choo2015} makes use of such a procedure for obtaining the luminance in a small simulated environment, structured from a carton box, a cheap web camera and a processing unit. On the other hand, Cai \etal \cite{cai2016luminance} instead focused on to a more advanced solution by taking advantage of the high dynamic range (HDR) computational photography and its corresponding high quality equipment. However, all the previous mentioned solutions require a pre-calibration step of the camera sensor where pixel values are mapped according to a known luminance source. %As every camera is different though, and the light-to-pixel-value function depends on wavelength, aperture, exposure-time, pixel-location within the image, camera sensitivity settings, and individual camera-to-camera differences someone can easily understand that this is not a trivial procedure. 
However, the light amounts illuminating a surface simply are not reliably recoverable from a pixel-like array of radiance values, because they are the product of the irradiance, the surface reflectance, and  complex inter-reflections between all surfaces in the 3D scene. Untangling them is very challenging in the general case – as well explained in the pioneering works on \textit{``inverse rendering''} by Ravi Ramamoorthi and Hanrahan %at Stanford
\cite{ramamoorthi2001}, Steve Marschner %at Cornell 
\cite{marschner1997}, and others.

\noindent \textbf{Computer graphics.} 
Light modeling and understanding are also studied in computer graphics for the creation of photo-realistic renderings \cite{hughes2014computer,marschner2015fundamentals}. The forward and inverse light transport theory physically simulates the path of transmitted light in a 3D environment and models the image as an integration process. This formalization requires the 3D mesh, the material surfaces and many other physical properties \cite{Chandraker2011,Krivanek2014,veach1995optimally}. To this end, many light models are proposed for retrieving and rendering scenes with as much as possible lifelike illumination. The most well known model is radiosity \cite{Cohen1985,Goral1984}, popular for its simplicity and efficiency. Other more advanced and recent approaches are the instant radiosity \cite{Keller1997} with its bouncing energy and the Virtual Point Lights (VPL), photon mapping \cite{Jensen1996} and progressive photon-tracing \cite{Hachisuka2008} with the idea of tracing photons from a light source through the scene and store their hits on diffuse surfaces in a so-called photon map.

\noindent \textbf{Light design software.} 
Relux \cite{relux}, DIALux \cite{dialux} and AGi32 \cite{agi32} are commercial CAD-design modelling software products that are commonly used in the lighting design field, for measurement and evaluation of lighting solutions. These software require the information of the luminaire specifications by the  manufacturers, the material and photometric properties of objects in the scene that normally are retrieved from online libraries and finally the CAD model of the indoor environment. %while the geometrical structure needs to be know in advance. 
The light simulation process in all these software is based on the radiosity method, and different variants of it \cite{radiance}. Recently two software prototypes, HILITE and LiteMaker \cite{hilite,litemaker}, are currently being developed in an academic environment and they try to combine a physically based real-time rendering with an interactive lighting simulation for complex architectural environments. HILITE uses a many-light global-illumination solution baked in light maps, including glossy material \cite{Luksch2013,Luksch2014}, while LiteMaker combines a multi-resolution image filtering technique with an interactive, progressive photon-tracing algorithm \cite{Hachisuka2008}.

\noindent \textbf{Our proposal.} We bring together the best of image processing and light design software: By using RGBD images only and without user input as in lighting design software; we propose a per-pixel exact lighting intensity estimates using a radiosity model that accounts for realistic environments. By contrast, computer graphics approaches pose emphasis on photorealistic appearances, caring more for shadows, reflections, liquids and smoke, rather than the physical lighting measure as we are interested in this application.

\section{RGBD2lux approach}\label{sec:RGBD2Lux approach}

Here we describe the proposed indoor RGBD camera-aided light modeling system for real scenes. Our computational pipeline, as shown in Figure \ref{fig:teaser}, first extracts a partial description of the indoor space geometry (\ie a 3D mesh) and it computes a simple reflective model for every object in the scene (Sec. \ref{sec:camera_aided_radiosity}). Given these two outputs and knowing the position of the light sources, it is possible (Sec. \ref{sec:radiosity_model}) to estimate illuminance using the radiosity model that is enriched (Sec. \ref{sec:indoor_lighting}) to account for real luminaries responses and specific light sensitivity curves. 

\subsection{Camera-aided 3D and reflectance modelling} \label{sec:camera_aided_radiosity}

Most simulated environments used for light intensity estimation need the \emph{a priori} information of the CAD model and the material properties of the objects in the room. This is often hard to obtain in every lighting setup scenario and, when the CAD model is available, it is common to contain structural inaccuracies of the environment. To this end,  we  use a depth sensor for retrieving the coherent information of the indoor environment 3D structure without using a CAD model. The main goal is to reconstruct the surface from the point cloud, represent it as a pair of vertices and faces (also known as patches) and associate at each face a scalar albedo value. 

% \begin{figure*}[thbp]
% 	\begin{center}
% 	    \includegraphics[width=1\linewidth]{images/all_pipe_line_pt_cloud.png} %\hspace{.3cm}
% % 		\includegraphics[width=0.48\linewidth]{images/render_room1_2.PNG} \\
% % 		\textit{Room1}% \hspace{3cm} \textit{Room2}
% 	\end{center}
% 	\caption{(a) Initial point cloud; (b) mesh given by the Delaunay triangulation; (c) Laplacian smoothing; (d) Extracted faces from the mesh.} \label{fig:all_pipe_line_pt_cloud}
% % 	\vspace{-5pt}
% \end{figure*}

The main issues to deal with is that depth sensors provide a sparse and nosier point cloud from which it is difficult to obtain a complete CAD like representation. Thus, given an exemplar point cloud from the RGBD sensor, we first apply a denoising procedure to remove any outlier points, we then apply a mesh reconstruction solution based on Poisson surface reconstruction \cite{kazhdan2013} approach and finally we post process the mesh with a Laplacian smoothing filter. % before we use the extracted faces as the needed patches for the form factors computation. % from where we throw rays into the space.

Using instead the images from the RGBD sensor, we record a time lapse video of the scene undergoing light variations given by the activation of the different light sources (\ie luminaries) in the room. Then we select a subset of images with the LIT method \cite{Lit2017} for which a single light source only is active. We use such images to estimate the pixel-wise albedo $\rho$ using a first-order spherical harmonics model \cite{basri2007photometric} with known surface normals obtained from the previous surface reconstruction step. %Even by using an approximated model, we compute a scalar reflectance values for each image pixel that experimentally provides satisfactory performance. 
Then these values are mapped to the surface faces as the mean value of the of the albedo related to each pixel which are falling within the area of each triangle.

% manually after we extracted and labeled each surface of our 3D model to individual materials and objects (we used online libraries likewise to the other commercial light planning software).

Given the partial 3D surface of the indoor environment and the scalar reflectance %and the light source positioning and emitting intensity 
we can now apply the radiosity model in order to estimate the illuminance at each surface patch.

%The procedure for obtaining a CAD models from depth data provides a coarse mesh, thus we want to examine whether it is sufficient for generating reliable lighting measures as shown in the experimental section. This is simply done by solving the same least-squares estimate of the radiosity model in order to obtain the values $\vec r$ from Eq. \eqref{eq:radiosity_matrix}. 

%Lastly another aspect is that the surface we obtain in this case it is likely to be open since the depth camera observes a partial portion of the room and certainly it cannot perceive the structure of the ceiling. This partial 3D model is not usable by most commercial lighting software, which request a complete closed 3D model instead. Our radiosity  model can simply deal with this problem by applying a closure procedure \cite{vanLeersum1989} where Leersum’s rectification spreads the closure adjustments over all of the calculated view factors matrix values. After forming (\ref{eq:radiosity_matrix}), solving the linear system provides the light measurement at each patch.

\subsection{Light modelling - Radiosity} \label{sec:radiosity_model}

The radiosity model \cite{cohen1993rri,maitre2015photon} % due to its simplicity, efficiency as well as because it
is the current state-of-art model in all commercial lighting simulation software in the lighting field (Relux \cite{relux}, DIALux \cite{dialux}, AGi32 \cite{agi32}). % in contrast to  other approaches \cite{Hachisuka2008,Jensen1996,Keller1997,ritschel2009}
    
In more details, radiosity represents the light arriving at each point within a 3D scene by considering both direct light sources and inter-reflections. Given our 3D mesh discretized into a set of $n$ triangular faces (or patches), we aim to compute the radiosity $r_i$ at each patch $i = 1 \ldots n$. The scalar $r_i$ is measured in $[W/m^2]$ and it is given by the direct emission of the room active sources (\eg the room luminaires) plus the inter-reflection given by the other patches. We consider the information about the light source position in 3D and their luminous intensity as known since luminaries may hardly be moved after their installation. Thus, the patches corresponding to the light sources are assigned with an emitting intensity value equal to the luminous intensity of the luminaires.

Given the available information, we can write the expression for radiosity with linear equations as: %This is modelled in terms of linear equations: 
\vspace{-5pt}
\begin{equation}\label{eq:radiosity}
    r_{i} = e_{i} + \rho_{i}\sum_{j=1}^{n-1} f_{ij} r_{j},
    \vspace{-5pt}
\end{equation}
where $i$ and $j$ index the $n$ patches, $e_i$ is a scalar for the self-emittance of patch $i$ ($e_i=1$ for light source patches in the unit measure, 0 otherwise), $\rho_i$ is the isotropic reflectivity of patch $i$ and $f_{ij}$ is the form factor between patch $i$ and $j$.

Form factors encode two main aspects: 
\begin{itemize}
    \item \textbf{Visibility}, if two patches are visible one from each other, \ie this value is equal to zero if there is no line of sight between them;
    \item \textbf{Distance and orientation}, encoding how well two patches ``see each other'', \ie low values correspond to very far patches with an oblique line of sight, high values instead refers to close fronto-parallel patches.
\end{itemize}
They are further constrained to be strictly non-negative and satisfy the reciprocity relation $a_{i}f_{ij} = a_{j}f_{ji}$, where $a$ is the area of each patch (note that the reciprocity relation is not symmetric unless the patches have the same size). % (we describe in Section \ref{sec:form_factors} how to encode the scene geometry into the form factors.)

Eq. \eqref{eq:radiosity} can be expressed as a global model by stacking all the equations into a matrix (radiosity matrix) and thus obtaining a linear system $\mat F \: \vec r = \vec e$ such as:
%
% \vspace{-8pt}
\begin{equation}\label{eq:radiosity_matrix}
    \resizebox{.9\hsize}{!}{$
    \underbrace{ \begin{bmatrix}
   				1 - \rho_1 f_{11} & -\rho_1 f_{12} &  \cdots &  -\rho_1 f_{1n}      \\
   				-\rho_2 f_{21} &  1 - \rho_1 f_{22} & \cdots & -\rho_2 f_{2n}      \\
   				\vdots & \vdots & \ddots &  \\
   				-\rho_n f_{n1} & -\rho_n f_{n2} &   \cdots      & 1 - \rho_1 f_{nn}
   				\end{bmatrix} }_\text{$\mat F$} \underbrace{ \begin{pmatrix}
   				r_1\\
   				r_2\\
   				\vdots\\
   				r_n
   				\end{pmatrix}}_\text{$\vec r$}
   				= \underbrace{\begin{pmatrix}
   				e_{1}\\
   				e_{2}\\
   				\vdots\\
   				e_{n} 
   				\end{pmatrix}}_\text{$ \hspace{-20pt} = \hspace{15pt} \vec e$}$ } %= \\ = \mat F \vec r = \vec e$},
\end{equation}
%
% \begin{equation}\label{eq:radiosity_matrix}
%     \resizebox{.9\hsize}{!}{$
%     \begin{bmatrix}
%   				1 - \rho_1 F_{11} & -\rho_1 F_{12} &  \cdots &  -\rho_1 F_{1n}      \\
%   				-\rho_2 F_{21} &  1 - \rho_1 F_{22} & \cdots & -\rho_2 F_{2n}      \\
%   				\vdots & \vdots & \ddots &  \\
%   				-\rho_n F_{n1} & -\rho_n F_{n2} &   \cdots      & 1 - \rho_1 F_{nn}
%   				\end{bmatrix}
%   				\begin{bmatrix}
%   				B_1\\
%   				B_2\\
%   				\vdots\\
%   				B_n
%   				\end{bmatrix}
%   				= \begin{bmatrix}
%   				E_{1}\\
%   				E_{2}\\
%   				\vdots\\
%   				E_{n} 
%   				\end{bmatrix}$ } %= \\ = \mat F \vec r = \vec e$}
% \end{equation}

% \begin{equation}\label{eq:radiosity_matrix}
%     \resizebox{.9\hsize}{!}{$
    
%   				\begin{bmatrix}
%   				B_1\\
%   				B_2\\
%   				\vdots\\
%   				{\color{red}B_i} \\
%   				\vdots\\
%   				B_n
%   				\end{bmatrix}
   				
%   				=
   				
%   				\begin{bmatrix}
%   				E_{1}\\
%   				E_{2}\\
%   				\vdots\\
%   				{\color{red}E_{i}}\\
%   				\vdots\\
%   				E_{n} 
%   				\end{bmatrix}
   				
%   				+
   				
%   				\begin{bmatrix}
%   				\\
%   				\\
%   				\\
%   				\\
%   				{\color{red}\rho_i F_{i1} \hspace{5pt} \rho_i F_{i2} \hspace{5pt} \hdots \hspace{5pt} \rho_i F_{in}} 
%   				\\
%   				\\
%   				\\
%   				\\
%   				\end{bmatrix}
   				
%   				\begin{bmatrix}
%   				{\color{red}B_1}\\
%   				{\color{red}B_2}\\
%   				{\color{red}\vdots}\\
%   				{\color{red}\vdots}\\
%   				{\color{red}B_n}
%   				\end{bmatrix} $}
% \end{equation}
%\todo{Make the equation above normal text size.}
%
\noindent where $\mat F$ is a $n \times n$ square matrix describing the geometry of the whole room, the $n$-vector $\vec r$ contains the associated radiosities at each patch and the self-emission $\vec e$ contains non-zero values at patches corresponding to active light sources. Now, solving for $\vec r$ requires the knowledge of geometry $\mat F$, object reflectance $\rho_i$ and the luminous intensity of the light sources $\vec e$. %Notice also that there is a scale ambiguity on the rows of the matrix $\mat F$ related to the scalar $\rho_i$ and $e_i$ so if both are not exactly defined the estimation of radiosity will be clearly affected. For this reason, in order to estimate scale-free values of radiosity we need the exact specification of the lighting system (\ie the vector $\vec e$) and reflectance of the surface ($\rho$).

%Once the radiosity values are computed, they can be converted to illuminance considering the Lambertian assumption \cite{lambert1870photometrie} and the fact that reflections are scattered isotropically and thus the illumination intensity is not a function of the direction of the ray/beam:
    
%\begin{equation}
%    L_{(x,y,\theta,\phi)} = L_{(x,y)}\neq f(\theta,\phi)
%\end{equation}

%\noindent but instead will be constant for different sets of measuring equipment and directly propotional to the exitance intensity $E_{(x,y)}$ with a convertion factor of $k$:

%\begin{equation}
%    E_{(x,y)} = L_{(x,y)}\cdot k = L_{(x,y)}\cdot \frac{\pi}{\rho}
%\end{equation}

%\noindent where $E_{(x,y)}$ is the illuminance measured in lux and $L_{(x,y)}$ the radiance.

%\subsection{Form factors computation} \label{sec:form_factors}

%In order to solve Eq. \eqref{eq:radiosity_matrix} we need to provide the form factors $f_{ij}$. Among the available numerical approaches, we adopt the ray-tracing-based method introduced by Malley \cite{malley1988}, due its more general applicability and efficiency. 

In order to compute the form factors, we uniformly sample rays within the unit disc (\ie the orthogonal projection of the unit sphere), whereby each point on the unit disc defines the direction of a ray in the space. Following \cite{malley1988}, we compute $f_{ij}$ as the ratio:
\begin{equation} \label{eq:ratio}
    f_{ij} = \frac{m_j}{m_i},
\end{equation}
where $m_j$ stands for the number of rays emitted by patch $i$ that reaches patch $j$, and $m_i$ is the total number of rays emitted by facet $i$.

\begin{figure}[thbp]
	\begin{center}

		\subfloat[][{\footnotesize Monte Carlo}]{\includegraphics[width=0.30\linewidth]{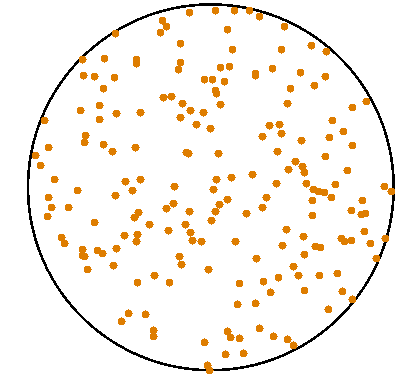}\label{fig:monte_carlo}}
		\hspace{0.01em}
		\subfloat[][{\footnotesize Isocell unit disc}]{\includegraphics[width=.26\linewidth]{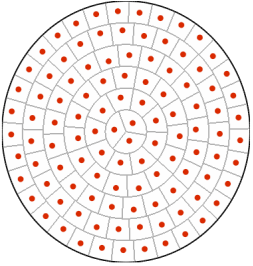}\label{fig:isocell}}
		\hspace{0.01em}
		\subfloat[][{\footnotesize Isocell unit sphere}]{\includegraphics[width=0.35\linewidth]{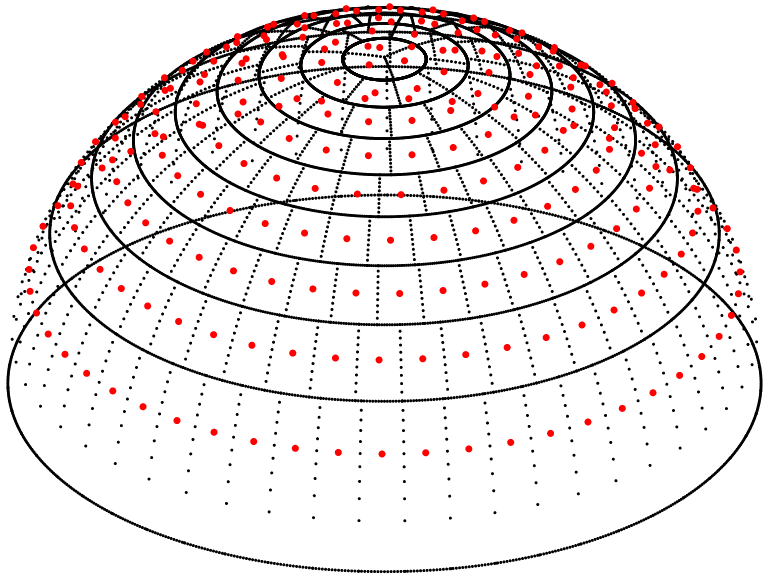}\label{fig:isocell_3d}}
% 		\hspace{0.01em}
% 		\subfloat[][{\footnotesize LDC and LSC curves}]{\includegraphics[width=0.2\linewidth]{images/ldc_lsc1.png}\label{ldc_lsc}}
        \vspace{5pt}
		\caption{Illustration of the two different ray distribution methods.} % \textcolor{blue}{TODO: if there is time replace first two subimages}}
		\label{fig:ray_methods}
	\end{center}
	\vspace{-12pt}
\end{figure}

As shown in Fig. \ref{fig:ray_methods}, there are different methods to distribute the rays \eg the Monte Carlo method \cite{Cook1984,Kajiya1986}, \ie the ray orientations are sampled randomly (see Fig. \ref{fig:monte_carlo}). This method always converges but requires a great number of rays to achieve a good precision, thus being computationally demanding. Instead we sample points in the unit disc (and therefore ray directions) by the approach of Masset \etal \cite{beckers2016, masset2011}, noted as ``Isocell'', due to its higher precision within a fixed computation time. This is based on a uniform discretization of the ray orientations (\cf Fig. \ref{fig:isocell}).
In our implementation, given our patch discretization and room size, we found that $1,000$ rays for each patch is a good compromise between accuracy and speed (\cf Fig. \ref{fig:isocel_rays} illustrates Isocell unit sphere rays, for a single patch on the office floor, within the CAD model).
Thus, for computing the matrix $f_{ij}$ we iterate through each patch and project the rays into the space from its center point and then we assign the corresponding form factor value according to the ratio described by Eq. (\ref{eq:ratio}).

Finally once we have populated the view factors matrix $f_{ij}$ we need to ensure the reciprocity relation mentioned earlier. Here we adopt the iterative scheme of Van Leersum \etal \cite{vanLeersum1989} that can be considered as a refinement of the naive form factors rectification.

\subsection{Customizing radiosity for real environments} \label{sec:indoor_lighting}

The radiosity model in Eq. \eqref{eq:radiosity_matrix} has two limitations: it assumes point light sources and it disregards the light perception. We address both aspects in our model extension, by introducing the LDC (for considering any light source) and the LSC (to model the light observer / sensor perception).

\begin{figure}[thbp]
	\begin{center}

		\subfloat[][]{\includegraphics[width=0.3\linewidth]{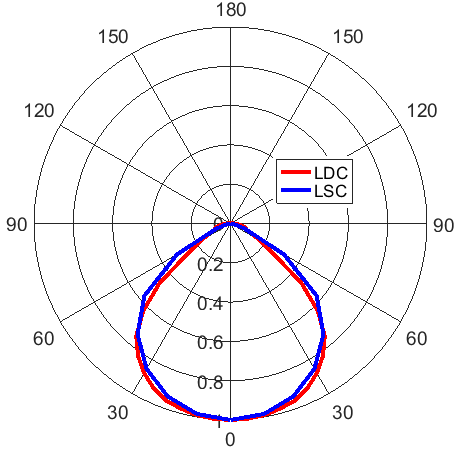}\label{ldc_lsc}\label{fig:ldc_lsc}}
		\hspace{0.01em}
		\subfloat[][]{\includegraphics[width=0.35\linewidth]{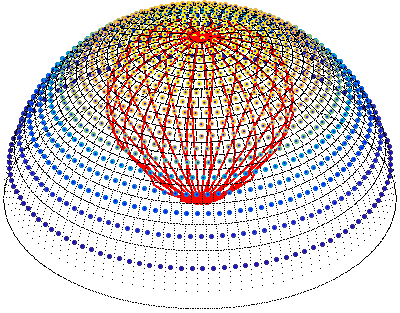}\label{fig:isocell_3d_weighted}}
		\hspace{0.01em}
		\subfloat[][]
		{\includegraphics[width=0.32\linewidth]{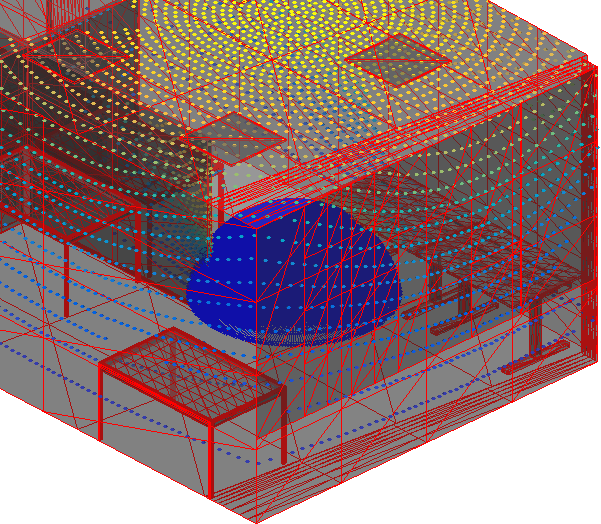}\label{fig:isocel_rays}}
        \vspace{5pt}
		\caption{(\textit{a}) Radial representation of LDC and LSC curves, illustrating how both quantities are attenuated (radius) with the growing light incident angle (radial angle); (\textit{b}) weighted ray distribution of the Isocell unit sphere for LDC, heatmap-color-coded; (\textit{c}) Isocell unit sphere (blue hemisphere) for a patch on the floor of the office CAD. The color-dots in (\textit{c}) are taken from (\textit{b}) but placed at the intersection of the cast ray with the other scene patches.} % \textcolor{blue}{TODO: if there is time replace first two subimages}}
		\label{fig:weighted_rays}
	\end{center}
% 	\vspace{-15pt}
\end{figure}

\vspace{-25pt}
\paragraph{Light distribution curve (LDC).} Let us recall Equations \eqref{eq:radiosity} and \eqref{eq:radiosity_matrix}. The radiosity model uses a scalar variable $e_i$ to set whether each patch is a light source ($e_i>0$) or not ($e_i=0$). The actual $e_i$ value would relate to the radiant intensity or luminous flux of the light source, measured in lumens $[cd/sr]$. Since it uses scalar values, the radiosity formulation of Section \ref{sec:radiosity_model} assumes isotropic light sources, \ie active patches radiate with the same intensity in all directions.

%If we are not interested to know the exact radiosity in the unit measure, we can assign a value of $1$ for an active patch and $0$ otherwise. 
%Likewise, the reflectivity value $\rho$ can be assigned at each patch in the CAD model. This procedure is  well automatized by light modelling software toolboxes \cite{relux,dialux} that provides a reflectance value given different furniture and surface in residential and office like environments. 

But isotropy is hardly the case for real light sources which are normally represented by a radiation map, such as in Figure \ref{fig:ldc_lsc} (\textit{red}). Their distribution is in general non-linear and encoded in a light distribution curve which provides the radiant intensity with respect to distance and angle from the emitting source (\ie how much light is emitted in each direction). The LDC curves are obtained experimentally in lab environment conditions with the use of specific equipment, \ie goniophotometer, and provided in datasheets by the luminaire manufacturer.

Thus, similarly to the existing solutions we include the LDC curve into the ray tracking procedure of our application and therefore we encode the non-linear radiation of light sources into the form factors $f_{ij}$. In more detail, we associate scalar values to each light source patch cast ray, which are proportional to the angle of emission, as illustrated in Figure \ref{fig:isocell_3d_weighted}. These are then used to re-write Eq. \eqref{eq:ratio} as a weighted mean. Since this procedure encodes the non-linear LDC into the form factors, our re-formulation of radiosity preserves the linearity of Eq. \eqref{eq:radiosity_matrix}.

Note that, in this way, the non-linearity is embedded in the form factor computation process, which is already non-linear. This results therefore in a minimal impact on computation, since the system in Eq. \eqref{eq:radiosity_matrix} remains linear.

% \begin{equation} \label{eq:ratio_ldc}
%     m_j = \sum (\frac{m_i * m_{j_{LDC}}}{\sum (m_{i_{LDC}})}),
% \end{equation}

% where $LDC_j$ are the weighted values corresponding to the rays reaching to patch $j$ and $LDC_r$ are all the weigthed values emmiting from patch $i$

\paragraph{Light sensitivity curve (LSC).} People perceive light differently, depending on their orientation with respect to it % (whether they are frontally illuminated or from the side) 
and depending on the distance from it.
In a similar fashion, luxmeter sensors have different sensitivity to lighting, depending on the lighting angle and distance as well as to manufacturing characteristics. The LSC plot in Figure \ref{fig:ldc_lsc} (\textit{blue}) illustrates the  perception characteristic of the luxmeter which we adopt in order to meet the measuring requirements of the collected ground truth data. Note the strong similarities to LDC as in Figure \ref{fig:ldc_lsc} (\textit{red}).

Therefore, we correct the sensor's light perception by the LSC. Similarly to the LDC, we formulate the corresponding weighted Isocell unit sphere, as in Figure \ref{fig:isocell_3d_weighted}, and integrate the weights into the form factor calculation by modifying the ray casting. In this way, we maintain the radiosity linear formulation and alter the computation time minimally. It is worth mentioning here that interestingly enough this functionality is not applicable in any  commercial light planning software available in the market.

\subsection{Method resume}

Algorithm \ref{alg:RGBD2lux} resumes the steps of the RGBD2lux approach necessary that estimates the light intensity using the input from an RGBD sensor. % and aligns it with the key elements of the radiosity light modeling.

\begin{algorithm}
	From depth data compute a 3D mesh, see Sec. \ref{sec:camera_aided_radiosity} ~\; \nllabel{alg:RGBD2lux_depth}
	From a time-lapse video sequence, identify a subset of images with single light activations and run photometric stereo to find the albedo values $\rho$, see Sec. \ref{sec:camera_aided_radiosity} ~\;
	Compute the form-factors $f_{ij}$ given the 3D mesh model, see Sec. \ref{sec:radiosity_model} ~\;
	For the 3D patches related to light sources apply the LDC computation of form factors, see Sec. \ref{sec:indoor_lighting} ~\;
	For the 3D patches where perceived light is measured, apply the LSC computation of form factors, see Sec. \ref{sec:indoor_lighting} ~\;
	Solve for the radiosity matrix, Eq. \ref{eq:radiosity_matrix}, see Sec. \ref{sec:radiosity_model} ~\;
	\nonl\caption{RGBD2lux method.}
	\label{alg:RGBD2lux}
\end{algorithm}

Once the radiosity values are computed by the RGBD2lux approach, they can be converted to illuminance considering the Lambertian assumption \cite{lambert1870photometrie} and the fact that reflections are scattered isotropically and thus the illumination intensity is not a function of the direction of the ray/beam:
    
\begin{equation}
    L_{(x,y,\theta,\phi)} = L_{(x,y)}\neq f(\theta,\phi),
\end{equation}

\noindent but instead will be constant for different sets of measuring equipment and directly proportional to the exitance intensity $E_{(x,y)}$ with a conversion factor of $k$:

\begin{equation}
    E_{(x,y)} = L_{(x,y)}\cdot k = L_{(x,y)}\cdot \frac{\pi}{\rho}
\end{equation}

\noindent where $E_{(x,y)}$ is the illuminance measured in lux and $L_{(x,y)}$ the radiance.
\section{Experimental Evaluation} \label{sec:exps}

Experiments are organized as follows. Sec. \ref{sec:LMB} presents a novel dataset for light measurements. Sec. \ref{sec:luxcom} shows quantitative results while Sec. \ref{sec:CADexp} describes in more details the evaluation of our radiosity model against state of the art approaches. Finally, Sec. \ref{sec:DEPTHexp} reports real experiments with the RGBD data only.

\subsection{Light measurement benchmark} \label{sec:LMB}

\begin{figure}[!ht]
	\begin{center}
	    \includegraphics[width=1\linewidth]{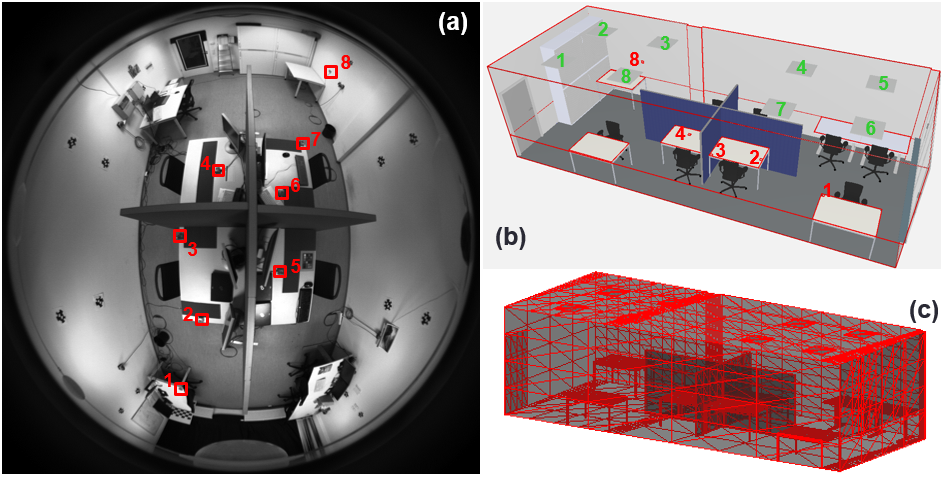} %\hspace{.3cm}
% 		\includegraphics[width=0.48\linewidth]{images/render_room1_2.PNG} \\
% 		\textit{Room1}% \hspace{3cm} \textit{Room2}
	\end{center}
	\caption{Room\_1 full scene. \textit{(a)} illustrate image as it looks from the camera, red bounding boxes are showing the location of luxmeters within the indoor space. Images \textit{(b)} and \textit{(c)} are detailed CAD design of the room showing the luminaire positioning and how the scene is subdivided into individual patches respectively.} \label{fig:rooms1_full_scene}
% 	\vspace{-5pt}
\end{figure}

To the best of our knowledge, there is currently no dataset for benchmarking light measurements with ground truth in real scenes.  Thus, we define a new benchmark for light measurement in real world scenes. We select two different office rooms, % \ie two office rooms, 
namely room\_1 and room\_2 (see Figure \ref{fig:rooms1_full_scene}). As shown, we provide a detailed CAD design for the rooms representing the 3D ground truth. This includes an accurate labelling of the object textures and reflectivities for each surface facet in the CAD.

Both rooms have been equipped with a controlled lighting environment, since the position of eight luminaries is known (Fig \ref{fig:rooms1_full_scene}b), as well as their type (Siteco Mira), meaning that the LDC curve is given by the manufacturer. Furthermore, to achieve even more accurate ground truth, we measured in lab conditions the overall luminous flux and temperature of each luminaire being $7913lm$ and $4250k$ respectively, while the multiplicative parameter for the luminaire age\footnote{The efficiency of every luminaire degrades over time.} is assigned to value $1$ %(needed for the Relux simulations) 
since the installation is fairly new.

We provide for both rooms a number of sensory data. First, we set up in the ceiling of the rooms a RGBD calibrated and aligned sensory system which consists of an rgb camera with fish-eye lens with \ang{180} FOV and a depth camera (Bluetechnix time-of-flight) with \ang{90} FOV. However, in our experiments we consider only the part that is visible to both images after we have registered and undistorted them (as shown in Fig. \ref{fig:teaser}, \textit{left}).  %applied an undistortion pre-processing.

The sensors are synchronized with luxmeters (also indicated in Fig. \ref{fig:rooms1_full_scene}a and \ref{fig:rooms1_full_scene}b), %measuring  accurate illumination in lux and
providing therefore the ground truth data for illumination intensity. The luxmeters however give localized (\ie point-to-point) lux readings only, so we installed 8 of them  in different areas so providing a reasonable sampling of the environment. We chose mainly locations over desk areas, since they are of major importance in the lighting field. For each luxmeter, we additionally report the type and their specific light sensitivity characteristic curve (LSC, see Fig. \ref{fig:ldc_lsc}), namely the sensor sensitivity across the incident light angles.

\begin{figure}[ht!]
	\begin{center}
		\includegraphics[width=1\linewidth]{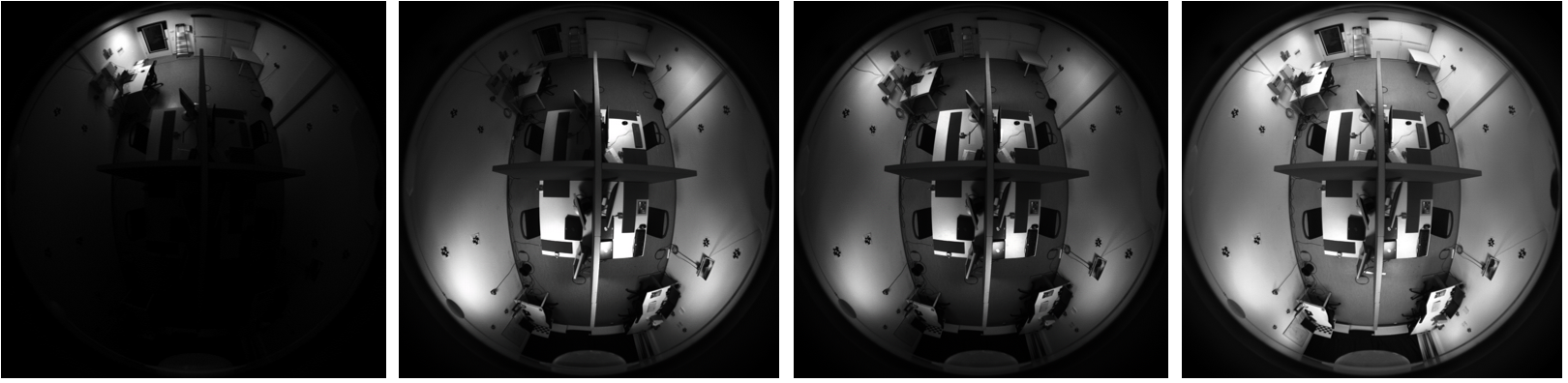}
	\end{center}
	\caption{Illustration of 4 illumination variants within room\_1. From the left to the right, the images illustrate the illumination provided by 1, 3, 4 and all 8 luminaires switched on in the scene.}\label{fig:illumination_combinations_room1}
% 	\vspace{-15pt}
\end{figure}

Thereafter, we evaluate $31$ %different scenarios with 
different luminaire activations (luminaires switched on or off) for each room, see Fig. \ref{fig:illumination_combinations_room1} for a sample of  images obtained from room\_1. We target the use of rgb and depth input just for light measurement, the use of luxmeters as ground truth, and all other provided information for ablation studies.

\subsection{Quantitative comparisons in Lux}\label{sec:luxcom}

Table \ref{table:quantitative_room1_2} summarizes the overall performance of the commercial light modeling software Relux, compared to ours and both to the ground truth under all the different experimental inputs. Beyond the proposed RGBD2lux, which does not require the CAD model and integrates the LDC and LSC curves, we additionally compare Relux to our own model using the CAD input provided (\ie \textbf{Ours with CAD}). Results are reported for both room\_1 and room\_2.

Note that our radiosity model using the knowledge of the CAD model outperforms Relux, thanks to the induced LDC and LSC distribution curves. In fact, against ground truth, \textbf{Ours CAD (LDC\_LSC)} achieves an average luxmeter error (across all 1-8 luxmeters) of $36$ Lux for room\_1 and $70$ Lux for room\_2\footnote{The supplemental material describes which are the structural differences between the two rooms together with the reason why room\_2 is more complex in terms of light modelling than room\_1}. These errors are $9.2\%$ and $3.7\%$ better than Relux, for room\_1 and room\_2 respectively.

Our proposed RGBD2lux (modelling LDC and LSC but not requiring the CAD) is still comparable with  Relux (instead requiring the CAD). Here we only consider the luxmeters in the area visible from the depth camera (\textbf{Avg (2-7)}), \ie those for which RGBD2lux may compute a Lux value. For room\_1 we achieve an error of $61$ Lux, and for room\_2 we achieve an error of 99 Lux. Thus, in this case the average lux error percentage is respectively worse than Relux by $4.6\%$ for room\_1, and by $8.1\%$ for room\_2.

It is of interest to note that the inclusion of LSC provides a larger error reduction than LDC (rows 6 and 7 in the table), consistently across the two rooms. We interpret this as the way the form factors matrix is populated, \ie based on the ratio between the rays arriving at a patch by the overall emitted rays, where there is a higher dependency on the emitted rays rather than the received rays, %LDC being partly addressed in the Isocell ray-tracing and its distribution within the Isocell unit disc. By contract, LSC stands for modelling the sensor sensitivity and its dependency with the received rays, 
which would otherwise be equally strong from any angle.

% Please add the following required packages to your document preamble:
% \usepackage{multirow}
\begin{table*}[!ht]
\centering
\caption{Illumination estimation errors by the considered approaches, applied to both rooms. Values for individual luxmeters, cols 1-8, correspond to the average lux values estimated over the results of the 31 different lighting combinations. Avg. (1-8) corresponds to the total average values for all installed luxmeters. By contrast, Avg. (2-7) only considers those luxmeters which are visible from the RGBD camera, \ie within its field-of-view. The percentage values correspond to the lux average percentage error in regards to the ground truth.}
\label{table:quantitative_room1_2}
\resizebox{.999\textwidth}{!}{\begin{tabu}{|c|c|c|c|c|c|c|c|c|c|c|l|c|c|c|c|c|c|c|c|c|c|}
\cline{1-11} \cline{13-22}
\multirow{3}{*}{\textbf{}}                                                                   & \multicolumn{10}{c|}{\textbf{\begin{tabular}[c]{@{}c@{}}Room 1\\ Error (in Lux)\end{tabular}}}                                                                                                                                          &  & \multicolumn{10}{c|}{\textbf{\begin{tabular}[c]{@{}c@{}}Room 2\\ Error (in Lux)\end{tabular}}}                                                                                                                                         \\ \cline{2-11} \cline{13-22} 
                                                                                             & \multicolumn{8}{c|}{\textbf{Luxmeters}}                                                               & \multicolumn{2}{c|}{\textbf{}}                                                                                                  &  & \multicolumn{8}{c|}{\textbf{Luxmeters}}                                                               & \multicolumn{2}{c|}{\textbf{}}                                                                                                 \\ \cline{2-11} \cline{13-22} 
                                                                                             & \textbf{1} & \textbf{2} & \textbf{3} & \textbf{4} & \textbf{5} & \textbf{6} & \textbf{7} & \textbf{8} & \textbf{\begin{tabular}[c]{@{}c@{}}Avg.\\ (1-8)\end{tabular}}  & \textbf{\begin{tabular}[c]{@{}c@{}}Avg.\\ (2-7)\end{tabular}}  &  & \textbf{1} & \textbf{2} & \textbf{3} & \textbf{4} & \textbf{5} & \textbf{6} & \textbf{7} & \textbf{8} & \textbf{\begin{tabular}[c]{@{}c@{}}Avg.\\ (1-8)\end{tabular}} & \textbf{\begin{tabular}[c]{@{}c@{}}Avg.\\ (2-7)\end{tabular}}  \\ \cline{1-11} \cline{13-22} 
\textbf{Relux}                                                                               & 167        & 96         & 27         & 26         & 43         & 10         & 96         & 39         & \begin{tabular}[c]{@{}c@{}}63\\ (21.4\%)\end{tabular}          & \begin{tabular}[c]{@{}c@{}}50\\ (20.7\%)\end{tabular}          &  & 206        & 97         & 27         & 80         & 97         & 49         & 73         & 44         & \begin{tabular}[c]{@{}c@{}}84\\ (22.2\%)\end{tabular}         & \begin{tabular}[c]{@{}c@{}}71\\ (20.4\%)\end{tabular}          \\ \cline{1-11} \cline{13-22} 
\textbf{\begin{tabular}[c]{@{}c@{}}Ours with CAD\\ (no\_LDC\_LSC)\end{tabular}}              & 188        & 150        & 33         & 45         & 43         & 34         & 91         & 65         & \begin{tabular}[c]{@{}c@{}}81\\ (27.5\%)\end{tabular}          & \begin{tabular}[c]{@{}c@{}}66\\ (27.3\%)\end{tabular}          &  & 207        & 114        & 99         & 148        & 105        & 117        & 93         & 81         & \begin{tabular}[c]{@{}c@{}}120\\ (31.8\%)\end{tabular}        & \begin{tabular}[c]{@{}c@{}}112\\ (32.2\%)\end{tabular}         \\ \cline{1-11} \cline{13-22} 
\textbf{\begin{tabular}[c]{@{}c@{}}Ours with CAD\\ (LDC)\end{tabular}}                       & 199        & 152        & 29         & 41         & 40         & 33         & 95         & 57         & \begin{tabular}[c]{@{}c@{}}81\\ (27.5\%)\end{tabular}          & \begin{tabular}[c]{@{}c@{}}65\\ (26.9\%)\end{tabular}          &  & 213        & 117        & 82         & 125        & 97         & 97         & 86         & 63         & \begin{tabular}[c]{@{}c@{}}110\\ (29.1\%)\end{tabular}        & \begin{tabular}[c]{@{}c@{}}100\\ (28.8\%)\end{tabular}         \\ \cline{1-11} \cline{13-22} 
\textbf{\begin{tabular}[c]{@{}c@{}}Ours with CAD\\ (LSC)\end{tabular}}                       & 73         & 45         & 24         & 32         & 40         & 34         & 46         & 52         & \begin{tabular}[c]{@{}c@{}}43\\ (14.6\%)\end{tabular}          & \begin{tabular}[c]{@{}c@{}}37\\ (15.3\%)\end{tabular}          &  & 69         & 80         & 98         & 136        & 70         & 84         & 56         & 62         & \begin{tabular}[c]{@{}c@{}}82\\ (21.7\%)\end{tabular}         & \begin{tabular}[c]{@{}c@{}}87\\ (25.0\%)\end{tabular}          \\ \cline{1-11} \cline{13-22} 
\textbf{\begin{tabular}[c]{@{}c@{}}Ours with CAD\\ (LDC\_LSC)\end{tabular}}                  & 69         & 24         & 22         & 38         & 28         & 28         & 38         & 41         & \textbf{\begin{tabular}[c]{@{}c@{}}36\\ (12.2\%)\end{tabular}} & \textbf{\begin{tabular}[c]{@{}c@{}}30\\ (12.4\%)\end{tabular}} &  & 70         & 57         & 76         & 106        & 75         & 69         & 55         & 53         & \textbf{\begin{tabular}[c]{@{}c@{}}70\\ (18.5\%)\end{tabular}}  & \begin{tabular}[c]{@{}c@{}}73\\ (21.0\%)\end{tabular}          \\ \cline{1-11} \cline{13-22} 
\textbf{\begin{tabular}[c]{@{}c@{}}Ours with CAD\\ Camera visible\\ (LDC\_LSC)\end{tabular}} & -          & 64         & 28         & 20         & 17         & 22         & 52         & -          & -                                                              & \begin{tabular}[c]{@{}c@{}}34\\ (14.1\%)\end{tabular}          &  & -          & 54         & 36         & 59         & 101        & 69         & 54         & -          & -                                                             & \textbf{\begin{tabular}[c]{@{}c@{}}62\\ (17.8\%)\end{tabular}} \\[-2.5pt] \tabucline[1.5pt]{1-11} \tabucline[1.5pt]{13-22}
\textbf{\begin{tabular}[c]{@{}c@{}}Ours RGB2Lux\\ (LDC\_LSC)\end{tabular}}                   & -          & 53         & 41         & 67         & 68         & 40         & 98         & -          & -                                                              & \begin{tabular}[c]{@{}c@{}}61\\ (25.3\%)\end{tabular}          &  & -          & 98         & 90         & 85         & 136        & 108        & 77         & -          & -                                                             & \begin{tabular}[c]{@{}c@{}}99\\ (28.5\%)\end{tabular}          \\ \cline{1-11} \cline{13-22} 
\end{tabu}}
\vspace{-8pt}
\end{table*}

\subsection{Comparisons against Relux using CAD models}
\label{sec:CADexp}

Here we analyze in more details our system against the commercial Relux software \cite{relux}. For comparison, since Relux requires the CAD model, we also provide this information to our method. These experiments are a sanity check for our proposed approach (since Relux also uses radiosity) and we provide a detailed analysis on the contribution of LDC/LSC curves and the effect of considering the visible scene only.

%In the following, we detail the reasons why our system  performs better than the commercial Relux software \cite{relux} when the CAD model is available as input. We consider this as a sanity check (since Relux also uses radiosity) and we verify that our enriched model with the LDC and LSC curves reduce the estimation error against the ground truth readings from luxmeters. %For both systems, we consider the CAD as input since this information is needed for Relux to work (we would remove this input for the RGBD based evaluation in the next section).

\begin{figure}[!ht]
	\begin{center}
	    \includegraphics[width=1\linewidth]{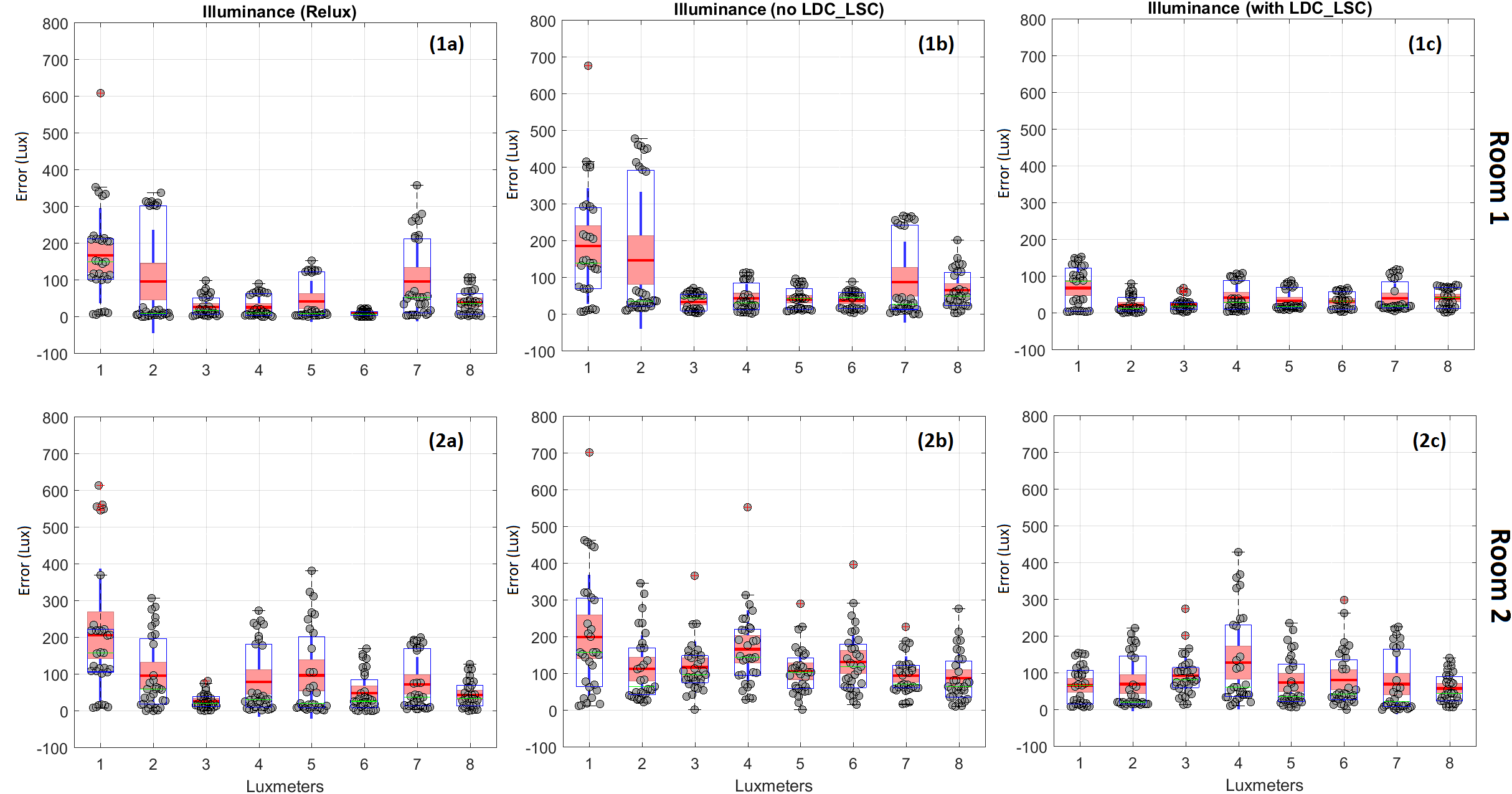} %\hspace{.3cm}
	   % \includegraphics[width=1\linewidth]{images/illuminance_d.png} %\hspace{.3cm}
% 		\includegraphics[width=0.48\linewidth]{images/render_room1_2.PNG} \\
% 		\textit{Room1}% \hspace{3cm} \textit{Room2}
	\end{center}
	\caption{Room\_1 \& 2 lux boxplot error evaluation using the full CAD information against \textit{(a)} Relux software, \textit{(b)} standard radiosity model and \textit{(c)} radiosity with \_LDC\_LSC applied.} \label{fig:rooms_full_scene_neg}
	\vspace{-5pt}
\end{figure}

Figure \ref{fig:rooms_full_scene_neg} shows six box plots  of the light measurement error (y-axis), as measured by each of the $8$ installed luxmeter sensors (x-axis). %as shown in Figure \todo{\ref{xxx}} (as discussed, each luxmeter represents the ground truth measurement).
In each column of the box plots, the $31$ gray dots represent the measured error of each of the lighting scenarios.
The pink box represents the central $50\%$ of the data, while the upper and lower vertical lines indicate the extension of the remaining error points outside it. The central red line indicates the mean error.

%\begin{figure}[!ht]
%	\begin{center}
%	    \includegraphics[width=1\linewidth]{images/illuminance_room1a.png} %\hspace{.3cm}
% 		\includegraphics[width=0.48\linewidth]{images/render_room1_2.PNG} \\
% 		\textit{Room1}% \hspace{3cm} \textit{Room2}
%	\end{center}
%	\caption{Room\_1 \& 2 lux error evaluation using the full CAD information against (a) Relux software, (b) standard radiosity model and (c) radiosity with \_LDC\_LSC applied.} \label{fig:rooms_full_scene}
% 	\vspace{-5pt}
%\end{figure}

%A global analysis of the results in Figure \ref{fig:rooms_full_scene_neg} shows that all the methods are under the median of $200$ lux error thus providing bounded results. This demonstrate as well that our customized radiosity model is at the level with state of the art solutions. 

In Figure \ref{fig:rooms_full_scene_neg}, we note a similar performance of Relux (leftmost box plots) and ours without LDC and LSC corrections, for both rooms (top and bottom rows), in line with the findings of Table \ref{table:quantitative_room1_2}.
Interestingly, highest errors occur at luxmeters 1, 2, and 7 which are closely localized under the luminaries and ``looking'' directly at the light source (check Figure \ref{fig:rooms1_full_scene} for  the sensors geometrical distribution). At these positions, the larger errors are due to a lower illuminance estimation by the radiosity model\footnote{See the signed plots in the supplementary material.} as it is also exemplified in Fig. \ref{fig:depth_illuminance_room1}.

\begin{figure}[!ht]
	\begin{center}
	    \includegraphics[width=1\linewidth]{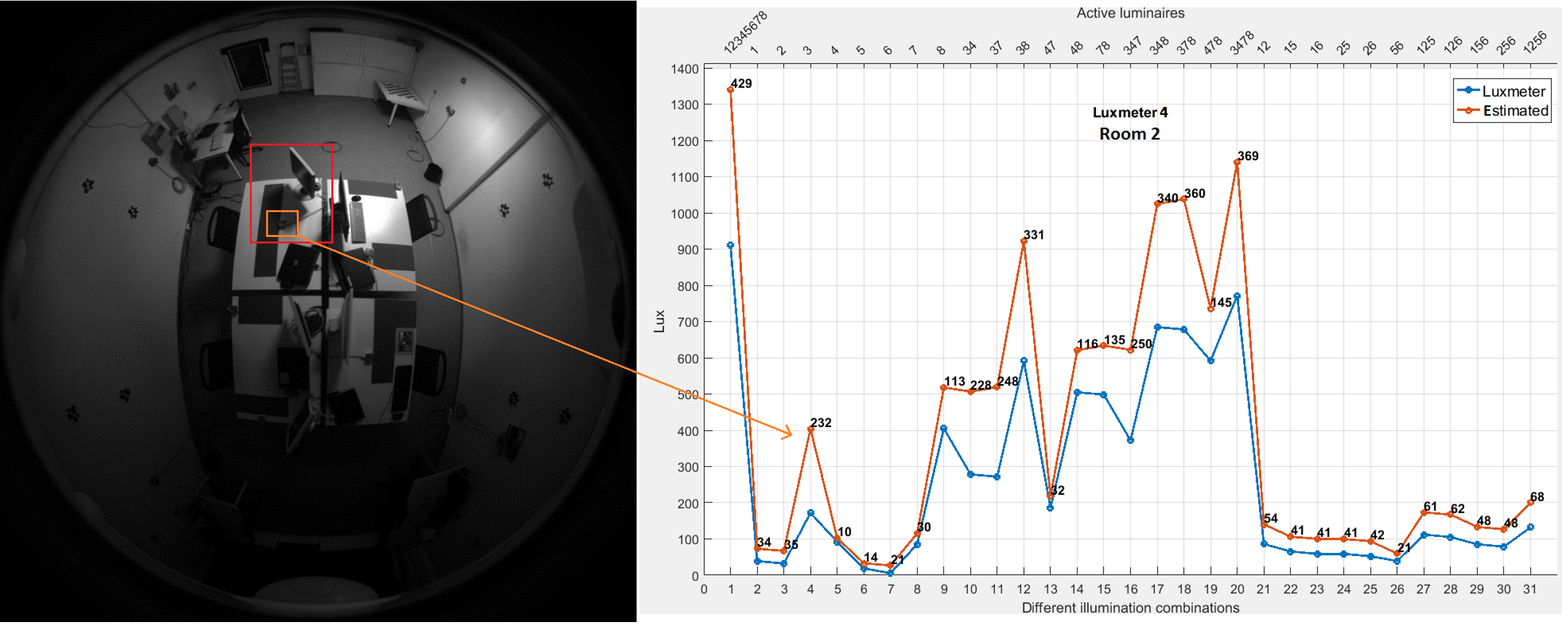}
	\end{center}
	\caption{A detailed description of the problem for luxmeter 4 in room 2. There are higher errors due to  shadows on the luxmeter sensor and inaccuracy in the 3D CAD model. The graph plot next shows the correlation of the sensor's measurement and which light activation the error is a result of. }\label{fig:error_room2}
% 	\vspace{-5pt}
\end{figure}

The rightmost plots in Figure \ref{fig:rooms_full_scene_neg} clearly show that the \_LDC\_LSC corrections are highly beneficial. The error reduces across all luxmeters and most prominently for those luxmeters 1, 2 and 7, addressing the orthogonally-incident light.
Of particular interest is the case of luxmeter 4 for room 2, whereby the error remains larger. Figure \ref{fig:error_room2} shows the region where the luxmeter is located and this reveals the cause of the issue. The screen monitor shields (occludes) luxmeter 4 from the light of some of the luminaires. Since the simplified CAD model does not include the monitor, the radiosity cannot model the occlusion properly. 

%  \begin{figure}[thbp]
%  	\begin{center}
%  	    \includegraphics[width=.9\linewidth]{images/mesh_partial_room_reprojected.png}
%  	\end{center}
%  	\caption{The scene partially represented from only the areas that are visible to the camera FOV.}\label{fig:partial_room}
% % % 	\vspace{-5pt}
%  \end{figure}

% \begin{figure}[!ht]
% 	\begin{center}
% 	    \includegraphics[width=1\linewidth]{images/illuminance_room1_2_intersections.png}
% 	\end{center}
% 	\caption{ Room1 \& 2 Lux error evaluation using the visible CAD information for our method using radiosity with \_LDC\_LSC.}\label{fig:visible_scene}
% % 	\vspace{-5pt}
% \end{figure}

\begin{figure}[!ht]
	\begin{center}
	    \includegraphics[width=1\linewidth]{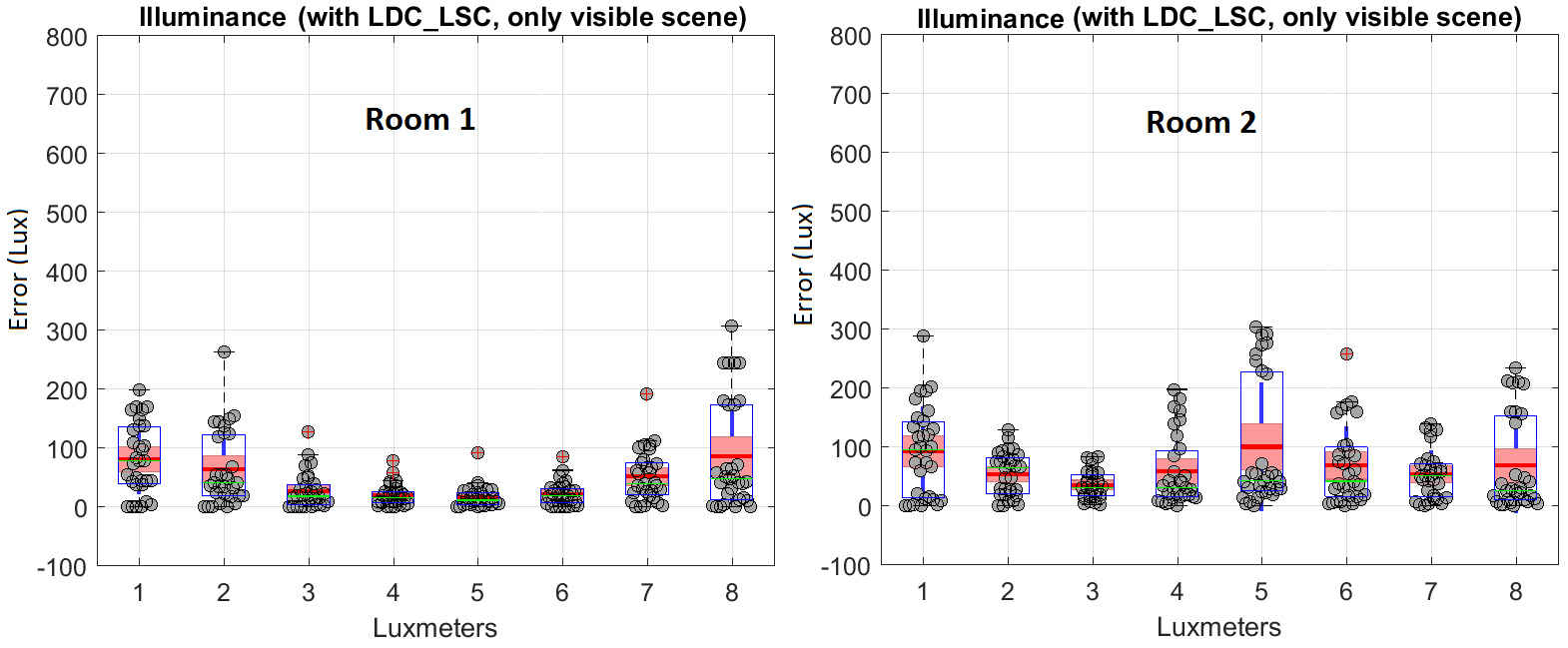}
	\end{center}
	\caption{ Room1 \& 2 Lux boxplot error evaluation using only the visible CAD information on the radiosity with \_LDC\_LSC.}\label{fig:visible_scene}
% 	\vspace{-5pt}
\end{figure}

Before we move to the depth based simulations we established another experiment where we simulated a partial part of the CAD model.
% based on it alignment to the real scene (as shown in Figure Fig. \ref{fig:partial_room}). 
Here we simulate that the only CAD information available is the one actually visible by the camera. Such information might be available for instance to a depth camera observing the scene. In such case almost $60\%$ of the room is not visible so providing a real challenge for our light estimation system. Figure \ref{fig:visible_scene} shows the results for our radiosity with LDC\_LSC model  since i) it is the best performing against standard radiosity, ii) Relux software does not work with open surfaces (while we do). 

%  \begin{figure}[!ht]
%  	\begin{center}
%  	    \includegraphics[width=.8\linewidth]{images/mesh_partial_room_reprojected.png}
%  	\end{center}
%  	\caption{{\small The scene partially represented from only the areas that are visible to the camera FOV.}}\label{fig:partial_room}
% % % 	\vspace{-5pt}
%  \end{figure}

Note the increment in the error for luxmeters 1, 2 and 8, especially in room 1, in comparison to Figure \ref{fig:rooms_full_scene_neg} \textit{(1c)}, which are closer to the walls. Since part of the walls are not visible (because they are not included in the camera field of view) the form factors fail to grasp the light contribution of the wall reflections (see Figure \ref{fig:visible_scene}).

\subsection{Light measurements from RGBD data} \label{sec:DEPTHexp}

Here we evaluate the performance of our proposed RGBD2lux. We use just the RGBD image and apply the correcting \_LDC and \_LSC curves to the radiosity model.

\begin{figure}[!ht]
	\begin{center}
	    \includegraphics[width=1\linewidth]{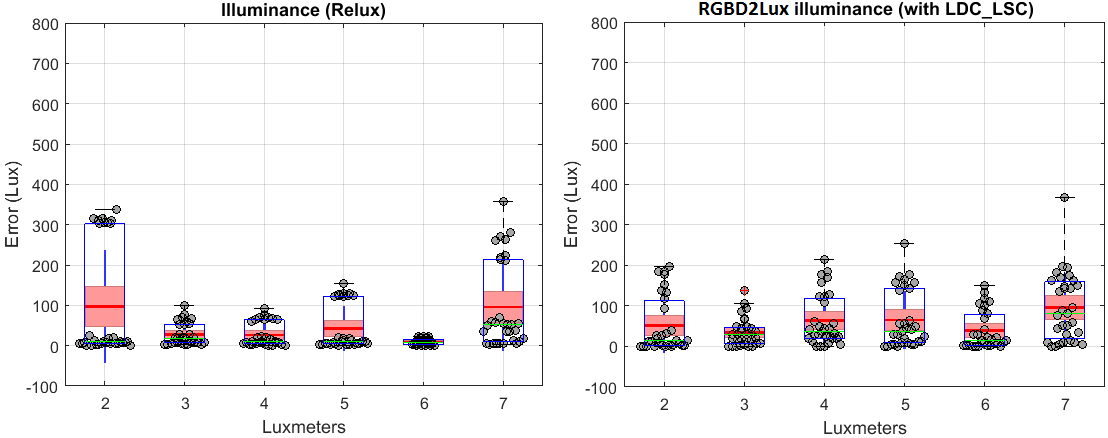}
	    \includegraphics[width=1\linewidth]{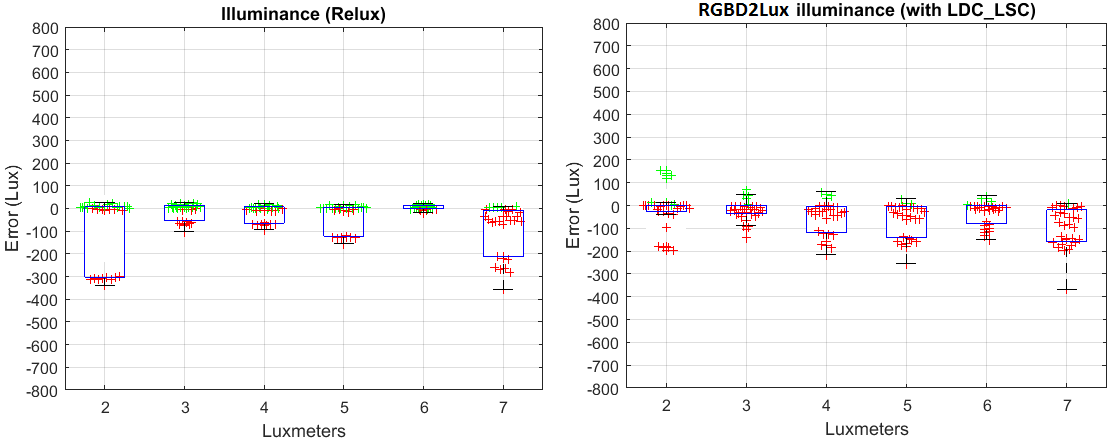}
	\end{center}
	\caption{The figure shows the comparison between the Relux simulation with CAD model \textit{(left)} and our RGBD2lux approach \textit{(right)} for room\_1. Even if our method do not use the CAD model, RGBD2lux achieves better or equivalent performance in some scenarios. Plots in the second row specify the type of the error, \ie due to lower \textit{(red marker)} or over \textit{(green marker)} estimation.}\label{fig:depth_illuminance_room1}
 	\vspace{-5pt}
\end{figure}

The box plots in Figure \ref{fig:depth_illuminance_room1} show that RBGD2lux achieves better or equivalent performance than the Relux simulation in some of the $31$ lighting scenarios tested. Note that the  results using the RGBD input have an error mainly due to lower estimation of the illumination levels within the room as a cause of the missing geometry. However, considering the incomplete geometry from the depth sensor, this experiment shows that our system can match or even overcome the CAD-based (with full geometry given) state of the art in challenging scenarios (similar results are obtained for room\_2 and included in the supplemental material). To conclude, the computational time for our modelling is approx. 3-5 minutes (Matlab) while Relux software simulations requires approx. 15-20 minutes for the same scenes.

\section{Conclusions}\label{sec:conclusion}
The proposed RGBD2lux method can be a viable option for estimating light intensity in real environments. Interestingly, we show that it is possible to challenge or even surpass current state-of-the-art light planning and modeling simulation software where the complete CAD model geometry is a necessary requirement. %, by using a partial geometry obtained from a camera sensor. 
Moreover, we show that the RGBD sensor, even if providing partial geometry information, can be a more realistic expression of the 3D structure of the current scene. Even the most accurate CAD model might not be aware about changes on the room structure that happens after the original planning. % Thus, to this end we answer to the question whether 3D modeling of a scene with a single camera is sufficient for reliable real scene light modeling and we set the baselines for adapting imaging and visual computing for smart lighting applications. % how light measurements in indoor environments can be made close to ground-truth, with a simple hardware setup consisting of an RGB-D camera, and no knowledge of the 3D structure of the room (CAD model).

This study provides a new direction for the research in lighting and for the lighting industry based on imaging and visual computing. On one side, we show that continuous and reliable camera-aided measurements of light on real environments is now viable, possibly  enabling the study of how light dynamics do influence human health, habits, social activities, to quote a few. 
The proposed lighting model can surely be improved and further experiments can be conducted, taking into account other material (\eg specular components, caustics, \etc) and scene (\eg dynamic scenes with people) properties or even go for a fully automatic solution where the positioning of the light sources will also be estimated. On the other side, lighting companies may design modules specifically suited for a fast and reliable light measurement. If attached to dimmable light sources, these modules may represent a major improvement for smart lighting, where light level is continuously updated to ensure maximum comfort, well-being and power efficiency as we successfully show in \cite{tsesmelis2018a}.
\newline
\newline
{\noindent\textbf{Acknowledgments:} The authors would like to thank Luc Masset for his insightful feedback. This project has received funding from the European Union's Horizon 2020 research and innovation programme under the Marie Sklodowska-Curie Grant Agreement No. 676455.}

{\small
\bibliographystyle{ieee}
\bibliography{egbib}
}

\end{document}